\documentclass{article}

\usepackage{arxiv}

\usepackage[utf8]{inputenc} 
\usepackage[T1]{fontenc}    
\usepackage{hyperref}       
\usepackage{url}            
\usepackage{booktabs}       
\usepackage{amsfonts}       
\usepackage{nicefrac}       
\usepackage{microtype}      
\usepackage{graphicx}
\usepackage{natbib}
\usepackage{doi}
\usepackage{bbding}
\usepackage{caption}
\usepackage{subcaption}

\newcommand{\tabincell}[2]{\begin{tabular}{@{}#1@{}}#2\end{tabular}}
\title{A Solution to CVPR'2023 AQTC Challenge: \\ Video Alignment for  Multi-Step Inference}

\date{} 					

\author{ {Chao Zhang$^{1,3}$, Shiwei Wu$^{2,3}$, Sirui Zhao$^{1,3}$, Tong Xu$^{1,2,3}$, Enhong Chen$^{1,2,3}$} \\
$^1$School of Computer Science and Technology, University of Science and Technology of China \\
$^2$School of Data Science, University of Science and Technology of China \\
$^3$State Key Laboratory of Cognitive Intelligence\\
	\texttt{zclfe00@gmail.com, \{dwustc, sirui\}@mail.ustc.edu.cn, \{tongxu, cheneh\}@ustc.edu.cn} \\
}

\renewcommand{\headeright}{Technical Report}
\renewcommand{\undertitle}{Technical Report}

\hypersetup{
pdftitle={A Solution to CVPR'2023 AQTC Challenge: Video Alignment for  Multi-Step Inference},
pdfsubject={q-bio.NC, q-bio.QM},
pdfauthor={David S.~Hippocampus, Elias D.~Striatum},
pdfkeywords={First keyword, Second keyword, More},
}

\begin{document}
\maketitle

\begin{abstract}
Affordance-centric Question-driven Task Completion~(AQTC) for Egocentric Assistant introduces a groundbreaking scenario. 
In this scenario, through learning instructional videos, AI assistants provide users with step-by-step guidance on operating devices.
In this paper, we present a solution for enhancing video alignment to improve multi-step inference.
Specifically, we first utilize VideoCLIP to generate video-script alignment features. 
Afterwards, we ground the question-relevant content in instructional videos.
Then, we reweight the multimodal context to emphasize prominent features.
Finally, we adopt GRU to conduct multi-step inference. Through comprehensive experiments, we demonstrate the effectiveness and superiority of our method, which secured the 2nd place in CVPR’2023 AQTC challenge. Our code is available at \url{https://github.com/zcfinal/LOVEU-CVPR23-AQTC}.
\end{abstract}

\section{Introduction}
AI assistants are progressively integrating into users' everyday routines.
To enhance the utilization of AI assistants, Affordance-centric Question-driven Task Completion~(AQTC) for Egocentric Assistant~\citep{wong2022assistq} has been introduced.
AQTC aims to assist users in navigating unfamiliar events for specific devices by providing step-by-step guidance using knowledge acquired from instructional videos.
Due to the substantial disparity among specific tasks, the integration of multimodal input, and the complexity of multi-step inference,
AQTC is still a challenging task.

Several studies have been proposed to address this task.
For instance, ~\citep{wong2022assistq} proposes Question-to-Actions (Q2A) Model, which employs vision transformer (ViT)~\citep{dosovitskiyimage} and BERT~\citep{kenton2019bert} to extract visual and textual features, respectively.
Moreover, attention mechanisms~\citep{bahdanau2015neural} are leveraged to anchor question-relevant information in instructional videos.
~\citep{wu2022winning} proposes a two-stage Function-centric approach, which segments both the script and video into function clips instead of sentences or frames.
Additionally, they substitute BERT with XL-Net~\citep{yang2019xlnet} for text encoding.
Despite the advancements achieved in AQTC through these techniques,
all of them adopt the unaligned pretrained encoders to extract visual and textual features, leading to significant semantic gaps between modalities~\citep{baltruvsaitis2018multimodal}, thereby hindering better results.

To alleviate the negative effects of modalities unalignment, in this paper, we leverage pretrained video-text models to achieve instructional video-text alignment, facilitating a more robust grounding of question-relevant knowledge for multi-step inference.
We build the pipeline with four steps: Instructional Video Alignment, Question-Aware Grounding, Multimodal Context Reweighting and Multi-Step Inference.
Specifically, we employ pretrained VideoCLIP~\citep{xu2021videoclip} for generating video-script alignment features, which are beneficial to cross-modal grounding.
Subsequently, we anchor the question-relevant content in instructional videos by the combination of hard and soft grounding.
Afterwards, we leverage additive attention~\citep{bahdanau2015neural} to adjust the weighting of the multimodal context to emphasize the salient features.
Finally, we employ GRU~\citep{chung2014empirical} for performing multi-step inference.
We reduce the proportion of teacher forcing linearly to bridge the gap between training and inference~\citep{zhang2019bridging}, which boosts the multi-step inference.

Our method attained the 2nd place in CVPR’2023 AQTC challenge.
Besides, we conduct exhaustive experiments to confirm the efficacy of our approach.

\section{Problem Definition}
In this section, we formulate the problem of AQTC.

Given an instructional video, which contains numerous frames and scripts, AI assistant extracts relevant information from the video in accordance with the user's question $q$.
Then, it deduces the correct answer $a^i_j$ based on the image $U$ as perceived by the user, from the candidate answer set $Ans^i=\{a^i_1,a^i_2,...,a^i_n\}$ in $i$-th step.
Following~\citep{wu2022winning}, we segment the video into several clips based on scripts.
Each clip illustrates one specific function of the device in video.
We concatenate these clips to form the visual function sequence as $[F_1^v,F_2^v,...,F_m^v]$ and the textual function sequence as $[F_1^t,F_2^t,...,F_m^t]$, where $F_i^v$ comprises all frames of the $i$-th function's clip, and $F_i^t$ contains all script sentences of the $i$-th function's clip.
To adapt AI assistant to the user's view, following~\citep{wong2022assistq}, we mask the referenced button related to candidate answers in user images $U$, denoted as $b_k$.

\section{Method}
\begin{figure}
	\centering
	\includegraphics[width=0.99\textwidth, height=0.636\textwidth]{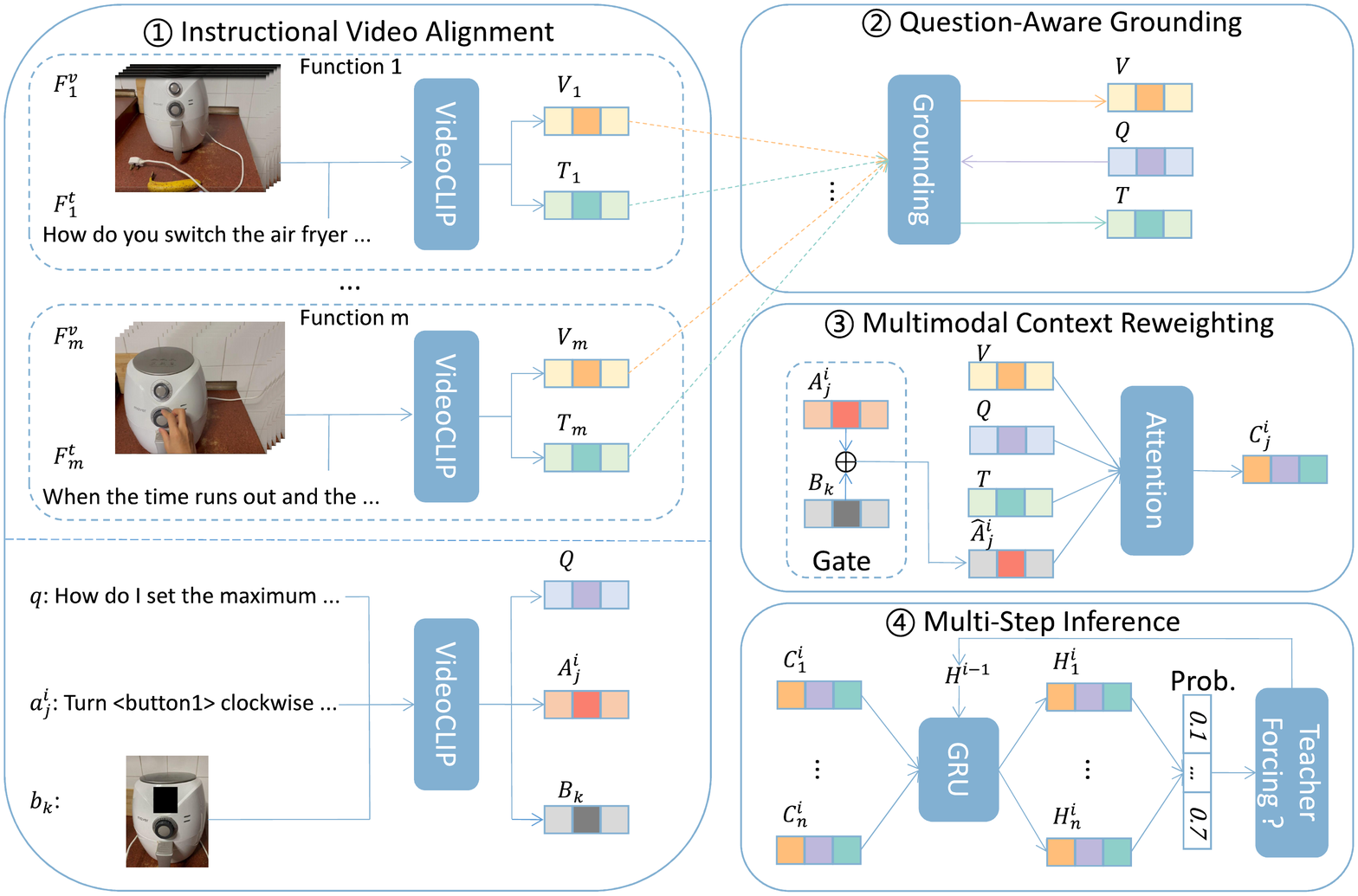}
	\caption{Framework of our method.}
	\label{framework}
\end{figure}
In this section, we will introduce the details of our method.
Our method consists of four steps: Instructional Video Alignment, Question-Aware Grounding, Multimodal Context Reweighting and Multi-Step Inference, as depicted in Figure~\ref{framework}.

\subsection{Instructional Video Alignment}
To align the videos and the text for better cross-modal understanding, we leverage pretrained VideoCLIP~\citep{xu2021videoclip} to generate the features of instructional videos. 
For the video part, we initially utilize pretrained S3D~\citep{miech19endtoend} to generate an embedding for each second of the video, with a frame rate of 30 frames per second.
Next, to represent each function within the videos, we utilize the pretrained visual transformer from VideoCLIP~\citep{xu2021videoclip} to process the embeddings generated by S3D~\citep{miech19endtoend} in each function.
Then, we apply average pooling over the processed sequence of embeddings to form the video embedding $V_i$ corresponding to a given visual function $F_i^v$.
For the text part, we use the pretrained textual transformer of VideoCLIP~\citep{xu2021videoclip} to encode the scripts of a textual function $F_i^t$.
Similarly, we employ average pooling to aggregate the processed sequence of text, generating the text embedding $T_i$ of a given textual function $F_i^t$.
Finally, we obtain the video feature sequence $[V_1,V_2,...,V_m]$ and the text feature sequence $[T_1,T_2,...,T_m]$ of the given function sequence.

Besides, we also utilize VideoCLIP~\citep{xu2021videoclip} to encode the questions $q$, the answer $a^i_j$ and the masked button image $b_k$.
We duplicate the images 30 times to ensure consistent video encoding.
We get the question feature $Q$, answer feature $A^i_j$ and visual button feature $B_k$.

\subsection{Question-Aware Grounding}
Owing to the extensive pretraining of VideoCLIP~\citep{xu2021videoclip} on a vast collection of videos, the features of videos and text are cross-modal aligned.
Therefore, we can utilize the question $Q$ to ground the video and text feature sequence directly.
Specifically, we leverage three grounding mechanisms: soft, hard and combined grounding.
Soft grounding employs attention~\citep{bahdanau2015neural} to learn the similarity between the question feature $Q$ and the video feature sequence $[V_1,V_2,...,V_m]$ directly.
And, it uses another attention network to compute the similarity between the question feature $Q$ and the text feature sequence $[T_1,T_2,...,T_m]$.
Soft grounding adopts the similarity from two attention networks to perform a weighted average of the two feature sequences, respectively.
Instead of relying on deep learning methods, hard grounding follows~\citep{wu2022winning}, which uses TF-IDF model~\citep{ramos2003using} to calculate the similarity between the question $q$ and each textual function $F_i^t$ from textual function sequence $[F_1^t,F_2^t,...,F_m^t]$.
Then, it uses the similarities as the weights to compute the averages of the video feature sequence $[V_1,V_2,...,V_m]$ and the text feature sequence $[T_1,T_2,...,T_m]$, respectively. 
Besides, the combined grounding utilizes soft grounding and hard grounding simultaneously.
Then, the two features from two grounding methods are averaged.
Ultimately, we obtain the aggregated question-aware video feature $V$ and text feature $T$.

\subsection{Multimodal Context Reweighting}
After obtaining multimodal question-aware context features from instructional videos, we need to model the answers to determine the correct one.
Specifically, we utilize the gate network~\citep{chung2014empirical} to fuse the candidate answer feature $A_j^i$ with the corresponding button feature $B_k$, which generates the multimodal answer feature $\hat{A}_j^i$.
We concatenate these multimodal contexts into a sequence $[V,T,Q,\hat{A}_j^i]$ for each candidate answer.
Due to the varying importance of different context features in determining the correct answers, we utilize additive attention~\citep{bahdanau2015neural} to reweight the multimodal context and get the fused feature.
Finally, the fused feature is processed using a two-layer MLP to obtain the candidate answer context feature $C_j^i$.

\subsection{Multi-Step Inference}
Owing to the requirement for multi-step guidance in order to respond to the given questions, it is essential for models to perform multi-step inference.
Following~\citep{wu2022winning}, we utilize GRU~\citep{chung2014empirical} to infer the current correct answer by incorporating historical knowledge.
Specifically, we feed the previous hidden state $H^{i-1}$ and the contextual features $C_j^i$ of candidate answers in $Ans^i$ into the GRU.
Then, the resulting current hidden state $H_{j}^i$ for each candidate answer in $Ans^i$ is utilized to predict the correct answer in the $i$-th step.
We adopt a two-layer MLP and the softmax function on the concatenated current hidden states $[H_1^i, H_2^i, ..., H_n^i]$ to generate the probability of the correct answer.
Cross entropy is used to compute the loss.
While previous works~\citep{wong2022assistq,wu2022winning} utilize the state of the ground truth as the historical state of the next step $H^{i}$.
This causes a huge gap between training and inference~\citep{zhang2019bridging}.
To bridge this gap, we reduce the reliance on teacher forcing linearly.
In other words, we choose the hidden state of the most probable answer predicted by models as the historical state of the next step $H^{i}$, when a sample is selected for autoregressive training.

\section{Experiments}
\subsection{Dataset and Implementation Details}
We use AssistQ train@22 and test@22 sets to train and validate.
And we test our model on the AssistQ test@23 dataset.

In our experiments, we use Adam optimizer~\citep{kingma2014adam} with a learning rate $10^{-4}$.
The batch size is set to $16$,  the maximum training epoch is 100, and we adopt early stopping.
We randomly select $5\%$ of the training data as the validation set.
We also incorporate semi-supervised learning (SSL) to extract information from the validation and test sets.
Specifically, we consider predictions for which the predicted probability of the correct answer by the models exceeds $0.9$ as pseudo-labels for SSL training.
We set the probability of teacher forcing to $max(0,1-0.05*current\_epoch)$, which linearly decreases over training epochs.

\subsection{Performance Evaluation}
\begin{table}
\begin{subtable}[t]{0.5\linewidth}
  \centering
  \begin{tabular}{l|cc}
    \toprule
    Methods     & R@1 (\%)     & R@3 (\%)  \\
    \midrule
    Q2A & 67.5  & 89.2     \\
    Question2Function & 62.6 & 87.5 \\
    Ours     & 75.4 & 91.8      \\
    Ours (Ensemble)    & 78.4       & 93.8  \\
    \bottomrule
  \end{tabular}
  \caption{Performance evaluation .}
  \label{tab:all}
    \end{subtable}
  \begin{subtable}[t]{0.5\linewidth}
\centering
  \begin{tabular}{l|cc}
    \toprule
    Methods     & R@1 (\%)     & R@3 (\%)  \\
    \midrule
    ViT+XL-Net & 63.9 & 86.6 \\
    VideoCLIP (Ours)     & 75.4 & 91.8      \\
    \bottomrule
  \end{tabular}
  \caption{Impact of pretrain features.}
  \label{tab:pretrain}
  \end{subtable}
  \caption{Performance evaluation and impact of pretrain features.}
\end{table}
We present the performance evaluation on CVPR’2023 AQTC challenge test dataset in Table~\ref{tab:all}.
We find that our method outperforms baseline methods (Q2A~\citep{wong2022assistq} and Question2Function~\citep{wu2022winning}).
This superiority can be attributed to our utilization of a video-text aligned pretrained encoder for feature extraction.
The aligned features are beneficial to multi-step inference.
Furthermore, our method exhibits improved performance when the results are ensembled.
\subsection{Ablation Study}
\paragraph{Pretrain Feature}

To validate the efficacy of video-text aligned features, we conduct the ablation study, which adopts ViT~\citep{dosovitskiyimage} for processing the visual features and XL-Net~\citep{yang2019xlnet} for processing the text features.
As shown in Table~\ref{tab:pretrain}, we observe that the performance of method that uses the unaligned features drops sharply.

\paragraph{Grounding Methods}
\begin{table}
  \centering
  \begin{tabular}{l|l|cc}
    \toprule
    \tabincell{l}{Text Grounding\\\quad Soft\quad Hard} & \tabincell{l}{Video Grounding\\\quad Soft\quad Hard}     & R@1 (\%)     & R@3 (\%)  \\
    \midrule
    \quad \Checkmark \quad\quad \Checkmark & \quad \Checkmark \quad\quad \XSolidBrush  & 75.4 & 91.8   \\
    \quad \XSolidBrush \quad\quad \Checkmark & \quad \Checkmark \quad\quad \XSolidBrush & 75.1 & 89.2 \\
    \quad \Checkmark \quad\quad \XSolidBrush & \quad \Checkmark \quad\quad \XSolidBrush & 73.8 & 90.5    \\
    \quad \Checkmark \quad\quad \Checkmark & \quad \Checkmark \quad\quad \Checkmark & 71.8 & 89.8    \\
    \quad \Checkmark \quad\quad \Checkmark & \quad \XSolidBrush \quad\quad \Checkmark & 75.1 & 89.8    \\
    \bottomrule
  \end{tabular}
  \caption{Impact of grounding methods.}
  \label{tab:grounding}
\end{table}
To validate the effectiveness of various grounding methods, we use different grounding techniques to train this model.
The result is presented in Table~\ref{tab:grounding}.
We find that the model achieves optimal performance when the text grounding leverages combined grounding and the video grounding utilizes soft grounding.

\paragraph{Reweighting Mechanism}
\begin{table}

\begin{subtable}[t]{0.5\linewidth}
  \centering
  \begin{tabular}{l|cc}
    \toprule
    Methods     & R@1 (\%)     & R@3 (\%)  \\
    \midrule
    Ours     & 75.4 & 91.8      \\
    w/o reweighting     & 72.1 & 89.5      \\
    w/o SSL     & 72.5 & 92.1      \\
    \bottomrule
  \end{tabular}
  \caption{Impact of the reweighting mechanism and SSL.}
  \label{tab:reweight}
\end{subtable}
\begin{subtable}[t]{0.5\linewidth}
  \centering
  \begin{tabular}{l|cc}
    \toprule
    Methods     & R@1 (\%)     & R@3 (\%)  \\
    \midrule
    Linear Decay (Ours)     & 75.4 & 91.8      \\
    AutoRegression     & 74.4 & 91.1      \\
    TeacherForcing     & 74.1 & 88.5 \\
    \bottomrule
  \end{tabular}
  \caption{Impact of multi-step inference strategies.}
  \label{tab:table}
\end{subtable}
\caption{Impact of the reweighting mechanism, SSL and multi-step inference strategies.}
\end{table}
We show the result of the model without attention reweighting in Table~\ref{tab:reweight}.
We observe a considerable decrease in performance for the model lacking attention reweighting.
This is because the attention reweighting can discern and prioritize the most informative features within complex multimodal contexts.

\paragraph{Multi-Step Inference}

We evaluate different multi-step inference strategies, as demonstrated in Table~\ref{tab:table}.
We find that the performance of TeacherForcing is inferior to that of the Linear Decay strategy, which is employed by our approach.
This is because TeacherForcing widens the gap between training and inference.
We also observe that Linear Decay outperforms AutoRegression.
This is because teacher forcing is beneficial in preventing models from accumulating mistakes during the early stages of training.

\paragraph{SSL}
The performance of the w/o SSL model exhibits a significant drop, as shown in Table~\ref{tab:reweight}.

\section{Conclusion}
In this paper, we present a solution aimed at enhancing video alignment to achieve more effective multi-step inference for the AQTC challenge.
We leverage VideoCLIP to generate alignment features between videos and scripts.
Subsequently, we identify and highlight question-relevant content within instructional videos.
To further improve the overall context, we assign weights to emphasize prominent features.
Lastly, we employ GRU for conducting multi-step inference.
Our method achieves the 2nd place in CVPR’2023 AQTC challenge.
Besides, we conduct exhaustive experiments to validate the effectiveness of our method.

\bibliographystyle{unsrtnat}
\bibliography{references}

\end{document}